# Value Function Approximation via Low Rank Models

Hao Yi Ong[1]

*Abstract*— We propose a novel value function approximation technique for Markov decision processes. We consider the problem of compactly representing the state-action value function using a low-rank and sparse matrix model. The problem is to decompose a matrix that encodes the true value function into low-rank and sparse components, and we achieve this using Robust Principal Component Analysis (PCA). Under minimal assumptions, this Robust PCA problem can be solved exactly via the Principal Component Pursuit convex optimization problem. We experiment the procedure on several examples and demonstrate that our method yields approximations essentially identical to the true function.

## I. INTRODUCTION

One way to solve Markov decision processes (MDPs) is to compute the state-action value function from which the optimal policy can be extracted. The value function can be represented as a matrix where each entry corresponds to the value for a state-action pair. For practical MDPs, data encoding the value function routinely lie in millions or even billions of dimensions. The ability to accurately represent this data on a compact basis would presumably have an impact on a wide area of disciplines that rely on stochastic decision making, including robotics, automated control, economics, and manufacturing.

We consider the problem of compactly approximating an MDP state-action value function. To alleviate the curse of dimensionality and scale,[1] we must leverage on the fact that the value functions have low intrinsic dimensionality. That is, they lie on some low-dimensional subspace [1], are sparse in some basis [2], or lie on some low-dimensional manifold [3], [4]. The foundation of our approach is similar to that in value function approximation in reinforcement learning (RL). In RL, researchers have employed a wide variety of basis function schemes to approximate value functions, most commonly radial basis functions and CMACs [5]. Implicit in these methods is the assumption that the value functions can be captured accurately by a small set of features; *i.e.*, an intrinsic assumption about low-dimensionality.

In our problem, we decompose a data matrix formed by the state-action values as a *low-rank* part plus a residual, which is not necessarily sparse (as we would like). The data matrix is thus modeled as a superposition of a low-rank component and a sparse component. This is posed as a matrix decomposition problem under the broader framework of *Robust Principal Component Analysis* (PCA). In general, accurate decomposition of a matrix is impossible; but the knowledge that the matrix has low rank radically changes this premise, making the search for solutions meaningful [6], [7]. As far as the author knows, this is a novel application of Robust PCA to value function approximation.

The rest of this paper is organized as follows. Section II introduces the mathematical formulation of our problem, followed by Section III, which presents an approach for Robust PCA. Section IV validates our approach through several numerical experiments, and Section V concludes with a few remarks on future work.

## II. VALUE FUNCTION DECOMPOSITION

We begin with a brief review on MDPs followed by an abstract definition of the decomposition problem.

### A. Markov decision process

In an MDP, an agent chooses action $a_t$ at time $t$ after observing state $s_t$. The agent then receives reward $r_t$, and the state evolves probabilistically based on the current state-action pair. The explicit assumption that the next state only depends on the current state-action pair is referred to as the Markov assumption. An MDP can be defined by the tuple $(S, A, T, R)$, where $S$ and $A$ are the sets of all possible states and actions, respectively, $T$ is a probabilistic transition function, and $R$ is a reward function. $T$ gives the probability of transitioning into state $s'$ from taking action $a$ at the current state $s$, and is often denoted $T(s, a, s')$. $R$ gives a scalar value indicating the immediate reward received for taking action $a$ at the current state $s$ and is denoted $R(s, a)$.

To solve an MDP, we compute a policy $\pi^\star$ that, if followed, maximizes the expected sum of immediate rewards from any given state. The optimal policy is related to the optimal state-action value function $Q^\star(s, a)$, which is the expected value when starting in state $s$, taking action $a$, and then following actions dictated by $\pi^\star$. Mathematically, it obeys the Bellman recursion

$$Q^\star(s,a) = R(s,a) + \sum_{s' \in S} T(s,a,s') \max_{a' \in A} Q^\star(s',a').$$

The state-action value function can be computed using a dynamic programming algorithm called value iteration. To obtain the optimal policy for state $s$, we compute

$$\pi^\star(s) = \operatorname*{argmax}_{a \in A} Q^\star(s,a).$$

---
[1]H. Y. Ong is with the Department of Aeronautics and Astronautics, Stanford University, Stanford, CA 94305, USA haoyi@stanford.edu
[1]We refer to either the complexity of algorithms that increases drastically as dimension increases, or to their performance that decreases sharply when scale goes up.

*B. Matrix decomposition*

Suppose matrix $M \in \mathbf{R}^{m \times n}$ encodes the state-action values of an MDP, where $m$ and $n$ are the cardinalities of the state and action spaces, respectively. Intuitively, this scheme leverages the correlation between action values close to each other. We approximate $M$ via the decomposition

$$M = L_0 + S_0,$$

where $L_0$ has low-rank and $S_0$ is sparse; here, both components are of arbitrary magnitude. We have no knowledge of the low-dimensional column and row space of $L_0$, or its dimensionality. Similarly, we do not know the locations and number of the nonzero entries of $S_0$. We wish to obtain $L_0$ and $S_0$, a low-rank plus sparse approximation of the true state-action value function. This is achieved via Robust PCA.

*C. Robust PCA*

Classical PCA [1], [8], [9] seeks the best (in an $\ell_2$ sense) rank-$k$ estimate of $L_0$ by solving

$$\begin{array}{ll} \text{minimize} & \|M - L\|_F^2 \\ \text{subject to} & \mathrm{rank}\, L \leq k, \end{array} \quad (1)$$

with variable $L$. Here, $\|\cdot\|_F$ denotes the Frobenius norm of a matrix, *i.e.*, the square root of the sum of the squares of the entries. This problem can be efficiently solved via the singular value decomposition (SVD) and enjoys a number of optimality properties when the noise $S_0$ is small and i.i.d. Gaussian.

While Robust PCA shares the same problem definition (1) as classical PCA, it does not have the same simplifying assumptions about the noise. Unlike the small noise in classical PCA, the entries in $S_0$ can have arbitrarily large magnitude, and their support is assumed to be sparse but unknown.[2]

## III. Approach

This section demonstrates how to cast the matrix decomposition problem as *Principal Component Pursuit* (PCP) and discusses our choice of algorithm to solve it.

*A. Principal Component Pursuit*

We obtain the value function components through the PCP estimate[3]

$$\begin{array}{ll} \text{minimize} & \|L\|_* + \lambda \|S\|_1 \\ \text{subject to} & L + S = M, \end{array} \quad (2)$$

which can be solved by *tractable* convex optimization. Here, $\|\cdot\|_* = \sum_i \sigma_i(\cdot)$ is the nuclear norm of a matrix; *i.e.*, the sum of its singular values. $\|\cdot\|_1$ denotes the $\ell_1$-norm of a matrix seen as a long vector in $\mathbf{R}^{mn}$. Assuming that the low-rank component $L_0$ is not sparse and that the sparsity pattern of the sparse component $S_0$ is selected uniformly at random, the simple PCP solution perfectly recovers the low-rank and the sparse components [12]. In particular, all that PCP requires about $L_0$ is that its singular vectors are not spiky; *i.e.*, the $\ell_\infty$-norm of any singular vector is not too large. To avoid any ambiguity, our model for $S_0$ is this: take an *arbitrary* matrix $S$ and set to zero its entries on a random set; this gives $S_0$.

*B. Identifiability of low-rank and sparse components*

To make the problem of matrix decomposition meaningful, we impose that the low-rank component $L_0$ is not sparse. We consider the general notion of incoherence introduced in [13] for the matrix completion problem; this assumption concerns the singular vectors of the low-rank component. We write the singular value decomposition of $L_0 \in \mathbf{R}^{m \times n}$ as

$$L_0 = U \Sigma V^T = \sum_{i=1}^{r} \sigma_i u_i v_i^T,$$

where $r$ is the rank of the matrix, $\sigma_1, \ldots, \sigma_r$ are the positive singular values, and $U = [u_1, \ldots, u_r]$, $V = [v_1, \ldots, v_r]$ are the matrices of the left- and right-singular vectors. The incoherence condition with parameter $\mu$ states that

$$\max_i \left\| U^T e_i \right\|_2^2 \leq \frac{\mu r}{m}, \quad \max_i \left\| V^T e_i \right\|_2^2 \leq \frac{\mu r}{n}, \quad (3)$$

and

$$\left\| UV^T \right\|_\infty \leq \sqrt{\frac{\mu r}{mn}}. \quad (4)$$

Above, we define $\|\cdot\|_\infty$, *i.e.*, the $\ell_\infty$ norm of a matrix seen as a long vector. As discussed in [6], [13], [14], the incoherence condition asserts that for small values of $\mu$, the singular vectors are not spread out; *i.e.*, not sparse.

Another issue is if the sparse matrix has low-rank. This will occur if, say, all the nonzero entries of $S$ occur in a column or in a few columns. Consider the case where the first column of $S_0$ is the opposite of that of $L_0$, and where all the other columns of $S_0$ vanish. Then it is clear that we would not be able to recover $L_0$ and $S_0$ since $M = L_0 + S_0$ would have a column space equal to or included in that of $L_0$. To avoid such situations, we will assume that the sparsity pattern of the sparse component is selected uniformly at random.

*C. Perfect recovery via PCP*

Surprisingly, the simple PCP solution perfectly recovers the low-rank and sparse components under the minimal assumptions above. Of course, we also require that the rank of the low-rank component is not too large, and that the sparse component is reasonably sparse. Below, $n_1 = \max\{m,n\}$ and $n_2 = \min\{m,n\}$.

**Theorem 1.** *Suppose $L_0$ is $m \times n$, obeys* (3) *and* (4), *and that the support set of $S_0$ is uniformly distributed among all sets of cardinality $z$. Then there is a numerical constant $c$ such that with probability at least $1 - c n_1^{-10}$ (over the choice of support of $S_0$), Principal Component Pursuit* (2)

---

[2] The unknown support of the errors makes the problem more difficult than the matrix completion problem that has recently been much studied [10], [11].

[3] Although the name naturally suggests an emphasis on obtaining the low-rank component, we are interested in both the low-rank and sparse components.

with $\lambda = 1/\sqrt{n_1}$ is exact; i.e., $L = L_0$ and $S = S_0$, provided that

$$rank\, L_0 \leq \rho_r n_2 \mu^{-1} (\log n_1)^{-2} \quad and \quad z \leq \rho_z mn.$$

Above, $\rho_r$ and $\rho_z$ are positive numerical constants.

In other words, matrices $L_0$ whose principal components are reasonably spread can be recovered with probability nearly one from arbitrary and completely unknown corruption patterns as long as these are randomly distributed. In fact, this works for large values of the rank; i.e., on the order of $n_2/(\log n_1)^2$ when $\mu$ is not too large.

Another remarkable property is that there is no tuning parameter in this algorithm. Under the assumption of Theorem 1, minimizing

$$\|L\|_* + \frac{1}{\sqrt{\max\{m,n\}}} \|S\|_1$$

always returns the correct answer; i.e., in Eq. (2) choose

$$\lambda = \frac{1}{\sqrt{\max\{m,n\}}}.$$

In fact, the proof of the theorem in [12] gives a whole range of correct $\lambda$ values, and this is a sufficiently simple value in that range.

### D. Algorithm

For small problem sizes, say $\max\{m,n\} < 100$, PCP can be performed using off-the-shelf tools such as interior point methods [15]. This was suggested for low-rank and sparse decomposition in [14]. However, despite their superior convergence rates, interior point methods are limited by the $O(m^6)$ complexity of computing a step direction. More sophisticated methods with better complexity and convergence rates include iterative thresholding methods using continuation techniques [16], [17], Bregman iterations [18], and Nesterov's optimal first-order algorithm for smooth and non-smooth minimization [19]–[21]. An *Accelerated Proximal Gradient* (APG) algorithm was suggested for low-rank and sparse decomposition in [22]. APG inherits the optimal $O(1/k^2)$ convergence rate for this class of problems, and empirical evidence suggests that it can solve the convex PCP problem at least 50 times faster than straightforward iterative thresholding.

Despite its good convergence guarantees, however, the practical performance of APG does not show good accuracy and convergence across a wide variety of problem settings [23]. In this paper, we choose to instead solve the convex PCP problem Eq. (1) using an augmented Lagrange multiplier (ALM) algorithm introduced in [23], [24]. [12] reports that ALM achieves much higher accuracy than APG, in fewer iterations, and that it works stably across a wide range of problem settings with no parameter tuning.

The ALM method operates on the *augmented Lagrangian*

$$L(L,S,Y) = \|L\|_* + \lambda \|S\|_1 + \langle Y, M-L-S \rangle + \frac{\mu}{2} \|M-L-S\|_F^2,$$

where $\langle \cdot \rangle$ denotes the standard trace inner product. A generic Lagrange multiplier algorithm would solve PCP by iteratively computing $(L^{(k)}, S^{(k)}) := \text{argmin}_{L,S}\, L(L,S,Y^{(k)})$, and then updating the Lagrange multiplier matrix via $Y^{(k+1)} := Y^{(k)} + \mu(M - L^{(k)} - S^{(k)})$ [25].

For our decomposition problem, we avoid solving a sequence of convex programs by recognizing that $\min_L L(L,S,Y)$ and $\min_S L(L,S,Y)$ both have very simple and efficient solutions [12]. Let $S_\tau : \mathbf{R} \to \mathbf{R}$ denote the shrinkage operator $S_\tau x = sign(x) \max(|x| - \tau, 0)$, and extend it to matrices by applying it to each element. It can be shown that

$$\underset{S}{\text{argmin}}\, L(L,S,Y) = S_{\lambda \mu^{-1}}(M - L + \mu^{-1} Y).$$

Similarly, for matrices X, let $D_\tau(X)$ denote the singular value thresholding operator given by $D_\tau(X) = U S_\tau(\Sigma) V^T$, where $X = U \Sigma V^T$ is any SVD. Again, we can show that

$$\underset{L}{\text{argmin}}\, L(L,S,Y) = D_{\mu^{-1}}(M - S + \mu^{-1} Y).$$

Thus, a more efficient strategy is to first minimize $L$ with respect to $L$ (fixing $S$), minimize $L$ with respect to $S$ (fixing $L$), and then finally update the Lagrange multipler matrix $Y$ based on the residual $M - L - S$. Algorithm 1 summarizes this strategy. We choose $\mu = mn/4\|M\|_1$, as suggested in [24], and terminate the algorithm when $\|M - L - S\|_F \leq \delta \|M\|_F$, with $\delta = 10^{-5}$.

---

**Algorithm 1** PCP by Alternating Directions [23], [24]

1: **initialize:** $S_0 = Y_0 = 0$, $\mu > 0$
2: **while** not converged **do**
3:     $L^{(k+1)} := D_{\mu^{-1}}(M - S^{(k)} + \mu^{-1} Y^{(k)})$
4:     $S^{(k+1)} := S_{\lambda \mu^{-1}}(M - L^{(k+1)} + \mu^{-1} Y^{(k)})$
5:     $Y^{(k+1)} := Y^{(k)} + \mu(M - L^{(k+1)} - S^{(k+1)})$
6: **output:** $L$, $S$

---

## IV. NUMERICAL EXPERIMENTS

To validate our approach, we apply it on two classical stochastic problems: the mountain car and inverted pendulum. The performance of our low-rank plus sparse models is evaluated against the true state-action value functions.

### A. Mountain car

Following the problem definition in [5], the car starts from the position-velocity pair $(x, \dot{x})$ and follows the dynamics

$$\dot{x} := \dot{x} + 0.001a - 0.0025 \cos(3x)$$
$$x := x + \dot{x},$$

where $a \in [-1, 1]$ is the acceleration input. The car can take on the state values $(x, \dot{x}) \in [-0.07, 0.07] \times [-1.2, 0.6]$. To incentivize getting to the top of the mountain at $x_0 = 0.5$, the reward function is defined

$$R(x) = \begin{cases} 10, & x \geq x_0 \\ -1, & \text{otherwise.} \end{cases}$$

## B. Inverted pendulum

In this problem, we are interested in balancing an inverted pendulum in its unsteady upright equilibrium position. The system is described by the angle-angular speed tuple $\left(\theta, \dot{\theta}\right)$, and its dynamics are

$$\theta := \theta + \dot{\theta} dt$$
$$\dot{\theta} := \dot{\theta} + \left(\sin\theta - \dot{\theta} + \tau\right) dt,$$

where $dt = 0.3$ is the time period between decisions and $\tau \in [-1, 1]$ is the torque input. The state space is $(-\pi, \pi] \times [-10, 10]$. The reward function penalizes control effort while favoring an upright pendulum angular position at 0°:

$$R(s, a) = \exp(\cos\theta - 1) - 0.1a^2.$$

## C. Discretization

The two test problems are continuous MDPs. To solve them via value iteration, their state and action spaces are discretized into fine grids, and their transitions are modeled using the multilinear interpolation [26]

$$T(s, a, s') = Prob(s' \mid s, a)$$
$$= \sum_{s'_c} Prob(s' \mid s'_c) Prob(s'_c \mid s, a),$$

where $s'_c$ is the continuous state evolved from taking action $a$ at state $s$. Here, $Prob(s' \mid s'_c)$ is specified using multilinear interpolation, and $Prob(s'_c \mid s, a)$ is the problem dynamics. In our problems, $s'_c$ is deterministically computed from the state-action pair $(s, a)$, so $T$ reduces to

$$T(s, a, s') = Prob(s' \mid s'_c).$$

The discretization scheme is summarized in Table I.

TABLE I: Discretization scheme

| Problem | Variable | Min. | Max. | No. of values |
|---|---|---|---|---|
| Mountain car | $x$ | -0.07 | 0.07 | 50 |
| | $\dot{x}$ | -1.2 | 0.6 | 50 |
| | $a$ | -1 | 1 | 1000 |
| Inverted pendulum | $\theta$ | $-\pi$ | $\pi$ | 50 |
| | $\dot{\theta}$ | -10 | 10 | 50 |
| | $\tau$ | -1 | 1 | 1000 |

With this scheme, we can extract the low-rank plus sparse policy for some continuous state $s_c$ via

$$\hat{\pi}(s_c) = \underset{a \in A}{\operatorname{argmax}} \sum_s Prob(s \mid s_c) \hat{Q}(s, a).$$

Evaluating $\hat{Q}(s, a)$ involves finding the action corresponding to the column with the maximum value in the row corresponding to state $s$ in the matrix $M = L + S$. The lookup can be done efficiently in real-time by taking the relevant matrix-vector product for the singular vectors and values encoding $L$ and adding that to the appropriate sparse row in $S$.

## D. Evaluation criteria

To evaluate our controllers, we run 1,000,000 simulations on both MDPs and compare the performance of the optimal and low-rank plus sparse model policies.

*1) Mountain car:* The evaluation metric for the controllers is how long it takes to reach the mountain top given a randomly and uniformly generated initial configuration.

*2) Inverted pendulum:* The metric for the inverted pendulum controllers is how well the controller can keep the pendulum in the upright position. This metric is captured by the average Euclidean distance between the pendulum angular position and the upright position. The initial states are drawn uniformly at random from the state-space.

## E. Implementation

All computation was carried out on a system with a dual-core Intel i7 processor, with clock speed 2.7 GHz and 8 GB of RAM, running Mac OS X. The ALM algorithm was implemented by the authors of [22] on a single thread in MATLAB and C, which interacted with MATLAB through a MEX interface [27]. The value iteration algorithm and simulation program were implemented in Julia [28]. All code can be found together with documentation integrated onto a Jupyter notebook at

https://github.com/haoyio/LowRankMDP.

## F. Results

Table II summarizes the results based on the evaluation criteria described above. Note that sparsity is defined as the fraction of zero elements over the total number of elements in the sparse component $S$. The results suggest that there is little if any difference between the optimal policy and the one produced by the low-rank and sparse model—they achieve virtually the same level of performance as their optimal counterparts. This is despite the low-rank models requiring less than 2% and 13% of the number of entries in the original matrices for the mountain car and inverted pendulum problems, respectively.

TABLE II: Summary simulation results

| Mountain car | time-to-goal | rank | sparsity | non-zero entries |
|---|---|---|---|---|
| optimal | 54.461 | 1000 | 0 | $2.5 \times 10^6$ |
| low-rank | 54.461 | 11 | 0.009 | $4.887 \times 10^4$ |

| Inv. pendulum | deviation | rank | sparsity | non-zero entries |
|---|---|---|---|---|
| optimal | 0.441 | 1000 | 0 | $2.5 \times 10^6$ |
| low-rank | 0.442 | 50 | 0.075 | $3.136 \times 10^5$ |

Figure 1 shows the policy heat maps for the mountain car and inverted pendulum MDPs. The color of any cell indicates the numerical value of the best control input given the state. There is a slight difference between the mountain car policy heat maps on the top left and right corners of the insets. On the other hand, visual inspection reveals little to no difference between the inverted pendulum policy heat maps. Intuitively, this comparison demonstrates why the

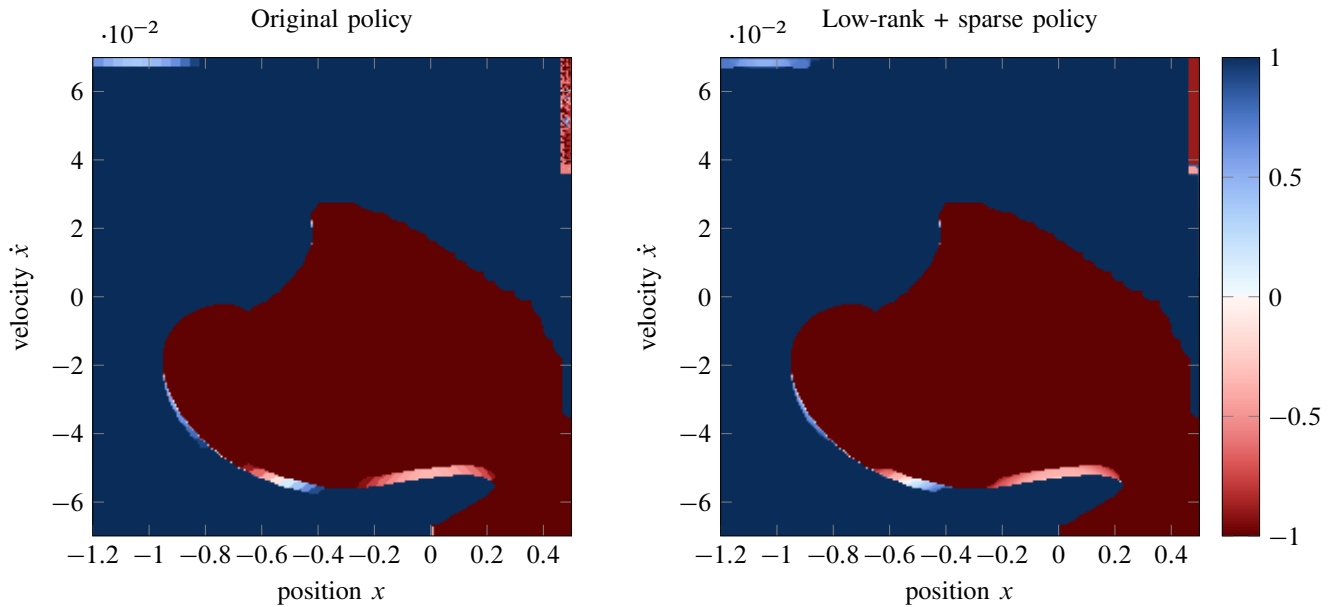

(a) Mountain car policy heat maps; notice the slight difference in the top left and right corners of the heat maps.

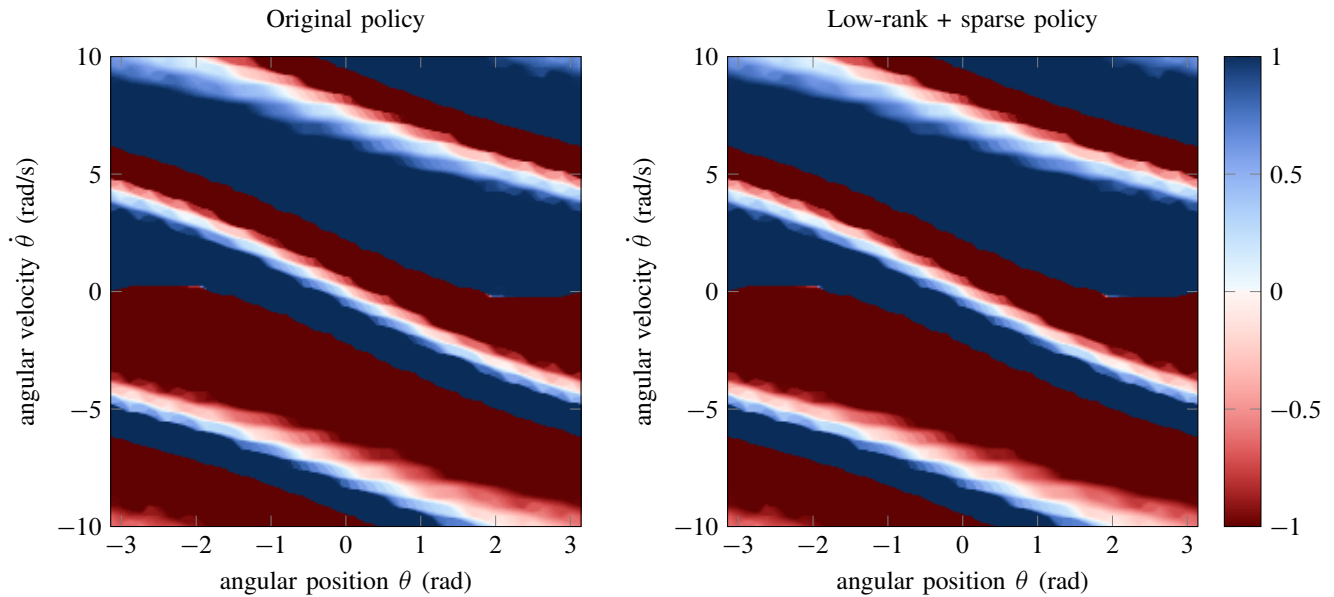

(b) Inverted pendulum policy heat maps; there is barely any difference between the two.

Fig. 1: Visual comparison of policy heat maps for the true and low-rank plus sparse models reveals little difference.

performance of the low-rank model was essentially identical to that of the original.

## V. CONCLUSION AND FUTURE WORK

We have demonstrated a novel value function approximation technique that exploits the intrinsic low dimensionality of MDPs. State-action value functions of simple continuous MDPs can be approximated virtually to perfection with far fewer memory requirements. It remains to experiment with a wider variety of MDPs to determine if the approach generalizes well. In the vein of applying Robust PCA to MDPs, an interesting research direction will be to frame reinforcement learning as a sequential noisy matrix completion problem.